\documentclass{article}

\usepackage{PRIMEarxiv}

\usepackage[utf8]{inputenc} 
\usepackage[T1]{fontenc}    
\usepackage{hyperref}       
\usepackage{url}            
\usepackage{booktabs}       
\usepackage{amsfonts}       
\usepackage{nicefrac}       
\usepackage{microtype}      
\usepackage{lipsum}
\usepackage{fancyhdr}       
\usepackage{graphicx}       
\graphicspath{{media/}}     

\usepackage{soul}
\usepackage[many]{tcolorbox}
\newcommand{\hlc}[2][yellow]{{%
    \colorlet{foo}{#1}%
    \sethlcolor{foo}\hl{#2}}%
    }
    
\title{Semantic Tokenizer for\\ Enhanced Language Processing
}

\author{
  Sandeep Mehta, Darpan Shah, R. Kulkarni \\
  Inspird, Inc. \\
  Mission Viejo, CA\\
  \texttt{\{smehta, dshah, rkulkarni\}@inspird.com} \\
\And
  Cornelia Caragea\\
  Computer Science \\
  University of Illinois at Chicago \\
  \texttt{cornelia@uic.edu} \\
}

\begin{document}
\maketitle

\begin{abstract}
Traditionally, NLP performance improvement has been focused on improving models and increasing the number of model parameters. NLP vocabulary construction has remained focused on maximizing the number of words represented through subword regularization. We present a novel tokenizer that uses semantics to drive vocabulary construction. 
The tokenizer includes a trainer that uses stemming to enhance subword formation. Further optimizations and adaptations are implemented to minimize the number of words that cannot be encoded. The encoder is updated to integrate with the trainer. The tokenizer is implemented as a drop-in replacement for the SentencePiece tokenizer.
The new tokenizer more than doubles the number of wordforms represented in the vocabulary. The enhanced vocabulary significantly improves NLP model convergence, and improves quality of word and sentence embeddings. Our experimental results show top performance on two Glue tasks using BERT-base, improving on models more than $50\times$ in size.
\end{abstract}


\section{Introduction}
NLP models have two primary components---a deep neural network and a vocabulary of embeddings. Recent improvements in NLP model performance have focused on improving deep networks and increasing model sizes. Interestingly, little attention has been paid to optimizing vocabularies. 

\begin{figure}[t]
 \center
 \includegraphics[width=\linewidth]{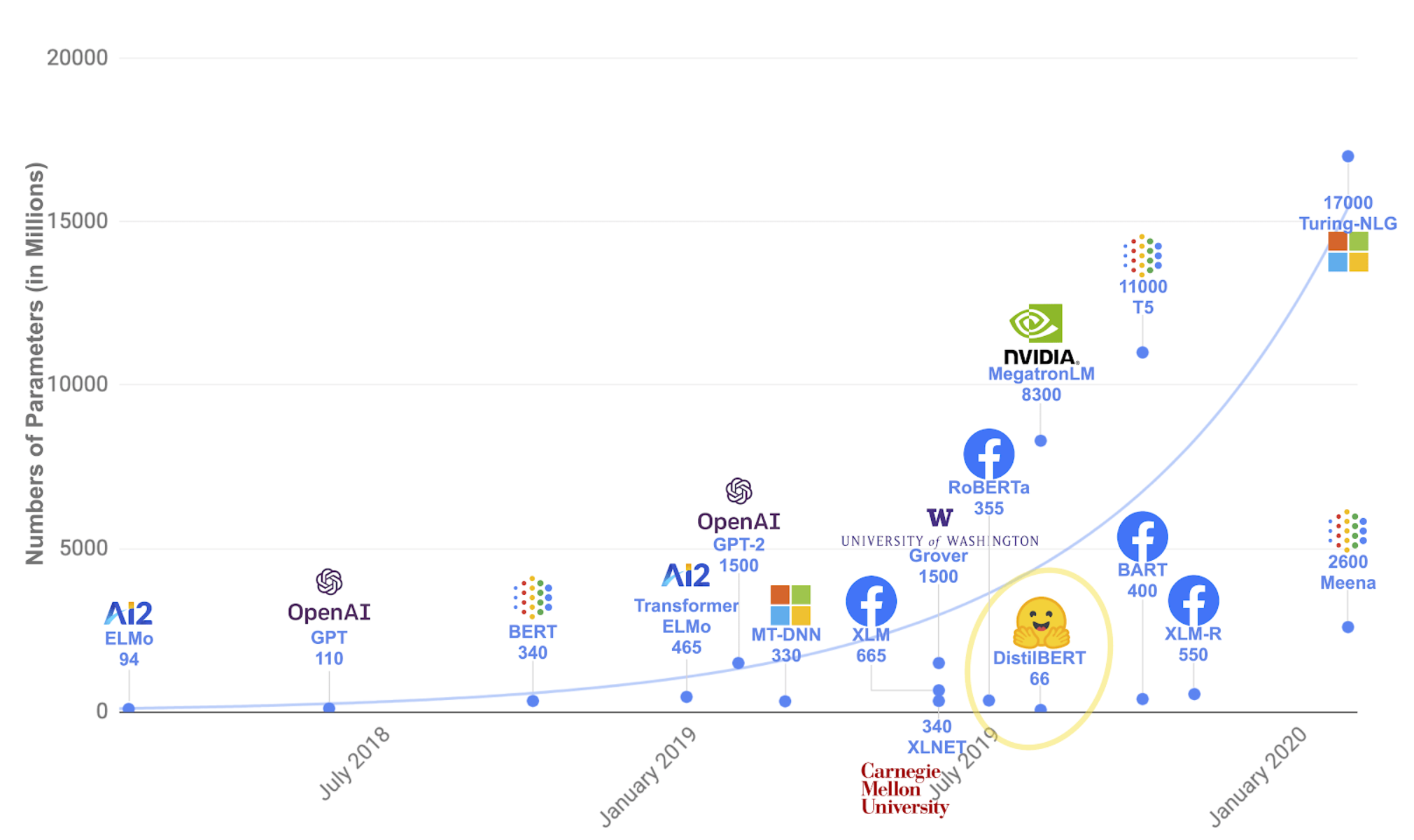}
 \caption{Increase in NLP Model Size.}
 \label{models}
\end{figure}

An analysis of recent models \cite{sanh2020distilbert}, plotted in Figure \ref{models},\footnote{Credit for figure to Huggingface’s DistilBERT: https://research.aimultiple.com/gpt/} shows that model sizes (i.e., number of parameters) have increased by 15,000\% over the last few years (excluding 175B parameter GPT-3 \cite{brown2020language} or 1T parameters of GPT4). The increase in model size significantly increases training costs. Recent publications  show that a single training run for GPT-3 could cost \$12M.\footnote{https://venturebeat.com/2020/06/01/ai-machine-learning-openai-gpt-3-size-isnt-everything/} Even BERT-Large \cite{devlin2018bert} training costs reach tens of thousands of dollars. Furthermore, increased model size place additional computational burden during model execution. These costs can have a detrimental effect on NLP innovation.

At the same time, the size of vocabularies has increased only about 100\% and the size of embedding vectors has increased about 200\%. Hence, over the same period, the fraction of NLP parameters representing the vocabulary has shrunk from 21\% in BERT-base \cite{devlin2018bert} to 0.3\% in GPT-3 \cite{brown2020language}. 

An average person uses 42,000 root words  and hundreds of thousands of wordforms \cite{Brysbaert2016HowMW}. Technical terminology and jargon add tens of thousands of additional words to the vocabulary. Even some of the largest vocabularies currently in use, DeBERTa \cite{deberta}, have only 128,000 tokens. Hence, NLP vocabularies need to include subwords that can be combined to form multiple words \cite{sennrich-etal-2016-neural}. Words that can not be represented using a single token are segmented into an initial subword followed by as many intermediate subwords as required. 

NLP models use a tokenizer to convert strings of characters into a sequence of lexical tokens. Tokenizers also construct the vocabulary of lexical tokens. We present a novel tokenizer that improves NLP performance by improving subword formation and embedding quality through {\em semantic tokenization}.

\section{Background}
\label{sec:background}
Tokenizers consist of three major components: 
\begin{enumerate}
    \item {\em Trainer:} The trainer builds the NLP vocabulary
    \vspace{-2mm}
    \item {\em Encoder:} Encoders are covert strings into a sequences of tokens
    \vspace{-2mm}
    \item {\em Decoder:} Decoders covert token sequences into words and sentences
\end{enumerate}

Commonly used tokenizers include WordPiece \cite{wordpiece}, Byte-Pair Encoding (BPE) \cite{sennrich-etal-2016-neural} and Unigram \cite{kudo-2018-subword}. Trainers in these tokenizer build  vocabulary through subword regularization. This subword regularization can be seen as an optimization function that maximizes the number of words in a corpus {\it D} that can be represented with a given vocabulary size $|V|$. Each tokenizer accomplishes this task with a different approach: BPE forms subwords by maximizing frequency of character sequences, WordPiece maximizes likelihood of subword formation and Unigram minimizes the loss. 

Several approaches are available for encoders as well. WordPiece uses a greedy longest-match-first strategy to tokenize a single word: it iteratively picks the longest subword of the remaining text that matches a word in the vocabulary. BPE encoder incrementally finds a set of subwords such that the total number of subwords for encoding the text is minimized. Unigram uses an entropy encoder that minimizes the total subwords for the text.

\section{Semantic Tokenizer}
\label{sec:tokenizer}
Subword regularization approaches have been shown to effectively model out-of-vocabulary words \cite{sennrich-etal-2016-neural}. This behavior is based on the intuition that many words, such as compound words, are formed from smaller common sub-units. Words that cannot be segmented into subwords generate [UNK] tokens which do not have any semantics. However, the subword regularization does not use any semantic information that may be readily available. 

An analysis of pretrained WordPiece/BERT \cite{devlin2018bert} and Unigram/ALBERT \cite{DBLP:conf/iclr/LanCGGSS20} vocabularies shows that about 75\% tokens represent initial subwords and 20\% are intermediate subwords. Most intermediate subwords do not have any inherent semantics. Many initial subword tokens are used to model different forms of the same root word. For example, the vocabularies contain eight forms of the word {\em advise}: {\em advise, advised, advisee, advisers, advises, advising, advisor}, and {\em advisors}. Furthermore, 15 related wordforms such as {\em advises, advisees, advisable} are not represented. Semantics quality decline rapidly when words are divided into more than 2 subwords.

We develop a new semantic tokenizer that enhances tokenization using semantics. Subsections below describe a new semantic tokenizer trainer and encoder.

\subsection{Trainer for the Semantic Tokenizer}
We reformulate subword regularization problem as a dual objective optimization problem: 
\begin{enumerate}
    \item {\em Maximize the embedding quality} 
    \vspace{-1mm}
    \item {\em Maximize the number of words that can be modeled} 
\end{enumerate}

To achieve these two objectives, we divide the the vocabulary $V$ into two segments $V_1$ and $V_2$. The segment $V_1$ generates subwords by focusing on semantics and maximizing embedding quality (Objective 1). The segment $V_2$ maximizes the number of words that can be modeled and minimizes the number of non-semantic [UNK] tokens (Objective 2). The parameter $f=\frac{|V_1|}{|V|}$ becomes part of the optimization problem and determines the size of the semantic segment of the vocabulary.

Using semantics to guide subword formation can be achieved by focusing on root words of each wordform.  Using suffixes as intermediate subwords can improve their semantic efficiency and free up more tokens in the vocabulary. Two linguistics morphological approaches are available for obtaining root words: Stemming and lemmatization. Stems are part of a word common to its inflected variants and lemmas are its canonical form, headword or root word. While lemmas are attractive because  they have inherent semantics in modeled language, stems are more attractive for use with subword formation. 

For example, forms of the word {\em advise} are represented with the stem {\em advis}. Different suffixes such as \#\#e, \#\#er, \#\#ing are used to form wordforms. Since stems are common across wordforms and suffixes are common across the vocabulary, the semantic trainer can represent many wordforms with a smaller set of stems and suffixes. Hence, stemming can better achieve the second tokenizer objective of minimizing number of subwords. Using same stem across wordforms improves semantic similarity between wordforms and improves input embedding quality. Furthermore, stemming improves NLP model training because increased number of occurrences of the stems and suffixes increases masking probability. In the example above, the stem {\em advis} occurs 230,000 times in the wiki corpus, while the word {\em advise} only occurs 10,159 times.  

The semantic trainer first populates the segment $V_1$. The trainer starts by generating a frequency table of the words in the training corpus and processes them in decreasing order of frequency. If stemming is possible for the word, it divides the word into its stem and suffix. If stem or the suffix has been previously been added, the code updates their likelihood. Otherwise, it adds the new stem or suffix to the vocabulary. If the word cannot be stemmed, it is added directly to the vocabulary. Once the first vocabulary segment is filled, the BPE algorithm is called to fill in the second segment $V_2$. The BPE algorithm is quite effective at achieving Objective 2 - minimize the number of [UNK] tokens.

We extended SentencePiece \cite{DBLP:conf/emnlp/KudoR18} with semantic training algorithms and a stemmer. Detailed analysis of vocabularies generated with available stemmers showed that the Snowball (Porter 2) stemmer \cite{porter2001snowball} is the most effective for use in the semantic trainer. The updated code provides a commandline parameter to set the relative size of segments. Experiments with wiki and book corpus showed that allocating 90 - 95\% to segment $V_1$ and 5-10\% to segment $V_2$ produced optimal results. 

\subsection{Encoder for the Semantic Tokenizer}
The encoder for the semantic tokenizer need to ensure usage of correct subwords in tokenization. When multiple subword sequences could represent a word, the encoder needs to select the sequence that uses the stem and suffix over any other non-semantic subword sequence. WordPiece's greedy longest-match-first strategy is directly applicable to semantic encoding. Hence, WordPiece encoder was used to without modification in the new tokenizer.

\section{Experimental Setup}

We used BERT-base model using the HuggingFace Transformers library \cite{wolf2019huggingface} for experiments. Since the focus of our work is to optimize the vocabulary, the benefits demonstrated are likely to scale to all NLP models with varied complexities.  

We used the semantic tokenzier trainer to build a new vocabulary and generated a version compatible with BERT. We generated training data using the new vocabulary, Wiki corpus and Book corpus. We trained the BERT Base model using a train batch size of 256; max sequence length of 256; max predictions per seq of 20; learning rate of 2e-4 and number of warmup steps = 10,000. The model was trained for 2M steps.

We compared performance of the new semantic tokenizer with pretrained BERT-base/Wordpiece and ALBERT/Unigram vocabularies.

\section{Results}

\subsection{Subword Regularization Efficiency}

We compared the new vocabulary against pretrained Wordpiece (BERT) and Unigram (ALBERT) vocabularies. We measured the tokenization efficiency by computing the number of words in Wiki Corpus and Book Corpus that can be represented by 1 or 2 subwords. As can be seen in Table \ref{tab:wordforms}, the new vocabulary covers a larger fraction of the words used in each corpus without using generic subwords. The new vocabulary is able to represent 125\% more wordforms or use 34\% fewer entries to represent the same number of wordforms. 

\begin{table}
\centering
\begin{tabular}{lccc}
\hline
\textbf{Corpus} & {\bf Unigram} & {\bf WordPiece} & {\bf Semantic}\\
\hline
{\bf Wiki} & 20,765 & 21,506 & 44,735 \\
{\bf Book} & 19,772 & 20,655 & 48,016 \\\hline
\end{tabular}
\caption{Number of Wordforms Represented.}
\label{tab:wordforms}
\end{table}

\begin{table}
\centering
\begin{tabular}{l|cccccc}
\hline
\multicolumn{1}{l|}{Corpus} & \multicolumn{2}{c}{\textbf{Unigram}} & \multicolumn{2}{c}{\textbf{WordPiece}} & \multicolumn{2}{c}{\textbf{Semantic}} \\
 \cline{2-7}
& Ave & CoV & Ave & CoV & Ave & CoV\\
\hline
Wiki & 1.9 & 47\% & 1.9 & 47\% & 1.7 & 51\% \\
Book & 2.0 & 44\% & 2.0 & 47\% & 2.0 & 48\% \\\hline
\end{tabular}
\caption{Subword Regularization Efficiency.}
\label{tab:vocab}
\end{table}

A concern with stemming is that it replaces a word with two subwords, which might increase the number of tokens and reduce the embedding quality. We evaluated number of tokens required to represent the corpus vocabulary and the variation in tokenization. As shown in Table \ref{tab:vocab}, this analysis showed that the semantic trainer actually reduced the average number of subwords per word.

\begin{table}
\centering
\small
\begin{tabular}{l|c|c|c}
\hline
Benchmark & BERT-base & BERT+Semantic & Leader \\
\hline
CoLA & 52.1 & \hlc[green!45!black!15] {\bf 77.9} & 74.4 \\
SST-2 & 93.5 & \hlc[yellow!35]{88.1} & 97.8 \\
MRPC & 88.9/84.5 & 85.8 / 84.3 & 93.9/91.8 \\
QQP & 71.2 & \hlc[green!45!black!15]{\bf 93.0 / 95.6} & 75.2/90.9 \\
MLNI-m & 84.6 & 82.0 & 91.9 \\
MLNI-mm & 84.6 & {\bf 88.0} & 91.4 \\
QNLI & 90.5 & 90.0 & 97.3 \\
RTE & 65.7 & {\bf 86.8} & 93.2 \\
WLNI & 56.3 & 56.3 & 95.9 \\\hline
\end{tabular}
\caption{Glue Benchmark Results,}
\label{tab:glue}
\end{table}

\subsection{Embedding Quality}

We evaluated the overall performance of the semantic tokenizer against Glue tasks (Table \ref{tab:glue}). The new tokenizer helped BERT-base outperform the leaders on the Glue leaderboard for CoLA and QQP tasks. On two additional tasks (MNLI-Mismatched and RTE), the new model was significantly better than BERT-base on which it is based. We hypothesize that semantic tokenization has significant advantages in single-sentence or sentence similarity tasks. 

Two additional tasks showed poorer performance (SST-2 and WLNI). Analysis of related datasets showed very long sequences. Our hypothesis is that the small max sequence length used in the BERT-base training has a significant disadvantage in processing long sequences. Considering the limitation of the model (small model, small max sequence length), this represents significant advantages of semantic tokenization.

\section{Analysis of Tokenizer Performance}
As shown above, the semantic tokenizer improved subword regularization efficiency and performed better on benchmarks. However, the root cause of this improvement was not clear. We devised two experiments to better understand the performance of the semantic tokenizer. First, We analyzed the quality of input embeddings generated by the semantic tokenizer with pretrained BERT WordPiece tokenizer. Next we evaluated the output embeddings generated by the BERT-bse with semantic tokenizer and pretrained BERT.

\subsection{Quality of Input Embeddings}
\begin{figure}[t]
    \center
    \includegraphics[scale=1]{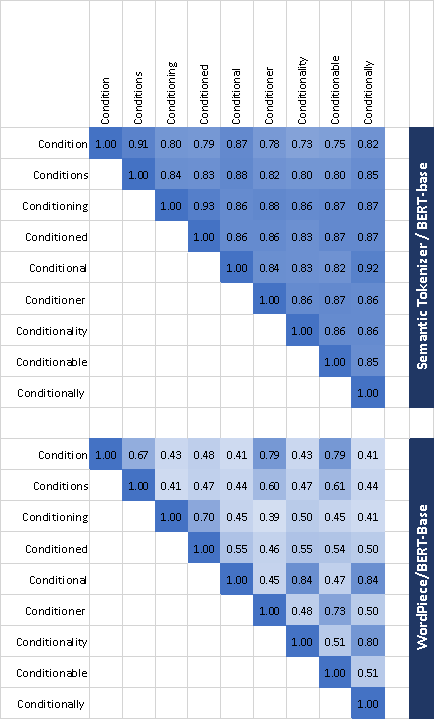}
    \caption{Wordform Embedding Similarities.}
  \label{cosineword}
\end{figure}

Input embeddings of different forms of a root word should be similar to ensure semantics are represented efficiently.  
We selected 9 forms of the word condition (Table \ref{tab:wordforms2}) and compared the new tokenizer embeddings with WordPiece/BERT. Figure \ref{cosineword} shows cosine similarities between embeddings across wordforms for each tokenizer. If a word was divided into subwords, we used mean pooling to compute its embedding. The cells are color coded to communicate the strength of similarity, with darker shading indicating higher similarity.

\begin{table}
\centering
\small
\begin{tabular}{llll}
\hline
\textbf{Word} & BERT/WP & BERT/Semantic\\
\hline
condition & condition & condit, ion \\
conditions & conditions & condit, ions\\
conditioning &conditioning & condit, ioning\\
conditioned & conditioned & condit, ioned\\
conditional & conditional & condit, ional\\
conditioner & condition, er & condit, ioner\\
conditionality & conditional, ity  & condit, ionality\\
conditionable & condition, able & condit, ionable\\
conditionally & condition, ally & condit, ionally\\
\hline
\end{tabular}
\caption{Tokenization of Test Wordforms}
\label{tab:wordforms2}
\end{table}

The new semantic tokenizer improved embedding similarity across different wordforms, despite using 40\% more tokens. On average, the semantic tokenizer improved embeddings by 40\% compared to WordPiece/BERT. Overall, these results prove that semantic tokenizer improves embeddings despite the increase in the number of tokens.

One explanation for this improvement is semantic concentration achieved through new subwords.  The stem {\em condit} occurs more than 300,000 times in the Wiki corpus. However the word {\em conditioner} occurs only 876 time.  Aggregation of stems and suffixes improves masking probability and hence embedding quality.

\subsection{Sentence Semantics}

\begin{figure}[t]
    \center
    \includegraphics[scale=1]{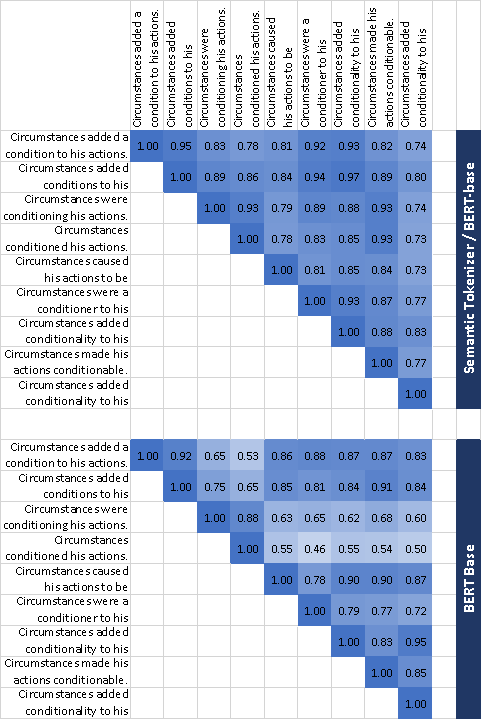}
    \caption{Sentence Embedding Similarity.}
  \label{cosinesent}
\end{figure}

We evaluated the impact of semantic tokenization on sentence level output embeddings. We composed 9 sentences with similar semantics using the 9 wordforms (Table \ref{tab:wordforms2}). We compared output embeddings of the semantic tokenizer with WordPiece/BERT-base in Figure \ref{cosinesent}. We computed the output embedding by mean pooling tokens at the output layer for each model. On average, cosine similarities across sentences improved 11\% compared to pretrained BERT-base/Wordpiece.  Overall, these results prove that stemming improves the accuracy of sentence level embeddings.

\section{Conclusion}
In this paper, we demonstrate significant NLP performance through improved vocabulary formation and without increasing model size. Specifically, we develop a new tokenizer that adds semantics as a constraint to vocabulary formulation. The tokenizer is implemented as a drop-in replacement for the SentencePiece tokenizer. The tokenizer includes a trainer that uses stemming to improve subword formation. An encoder is developed that complements the semantic trainer and integrates semantic tokenization into NLP models.
Overall, the new semantic tokenizer significantly improves NLP performance on processing, classification and search tasks without increasing model sizes. The new tokenizer more than doubles the number of wordforms represented in the vocabulary. The enhanced vocabulary significantly improves model convergence and quality of word embeddings. Our experimental using the tokenizer with BERT-base  show top performance on two Glue tasks while being 1/20th of the size of the other leading models.

\section*{Acknowledgments}
This work was performed in part under a US Air Force contract.

\bibliographystyle{unsrt}  
\bibliography{custom}

\begin{thebibliography}{10}

\bibitem{sanh2020distilbert}
Victor Sanh, Lysandre Debut, Julien Chaumond, and Thomas Wolf.
\newblock Distilbert, a distilled version of bert: smaller, faster, cheaper and
  lighter, 2020.

\bibitem{brown2020language}
Tom~B. Brown, Benjamin Mann, Nick Ryder, Melanie Subbiah, Jared Kaplan,
  Prafulla Dhariwal, Arvind Neelakantan, Pranav Shyam, Girish Sastry, Amanda
  Askell, Sandhini Agarwal, Ariel Herbert-Voss, Gretchen Krueger, Tom Henighan,
  Rewon Child, Aditya Ramesh, Daniel~M. Ziegler, Jeffrey Wu, Clemens Winter,
  Christopher Hesse, Mark Chen, Eric Sigler, Mateusz Litwin, Scott Gray,
  Benjamin Chess, Jack Clark, Christopher Berner, Sam McCandlish, Alec Radford,
  Ilya Sutskever, and Dario Amodei.
\newblock Language models are few-shot learners.
\newblock {\em Advances in Neural Information Processing Systems 33}, 2020.

\bibitem{devlin2018bert}
Jacob Devlin, Ming{-}Wei Chang, Kenton Lee, and Kristina Toutanova.
\newblock {BERT:} pre-training of deep bidirectional transformers for language
  understanding.
\newblock In {\em Proceedings of the 2019 Conference of the North American
  Chapter of the Association for Computational Linguistics: Human Language
  Technologies, {NAACL-HLT} 2019, Minneapolis, MN, USA, June 2-7, 2019, Volume
  1 (Long and Short Papers)}. Association for Computational Linguistics, 2019.

\bibitem{Brysbaert2016HowMW}
M.~Brysbaert, Micha{\"e}l~A. Stevens, Paweł Mandera, and Emmanuel Keuleers.
\newblock How many words do we know? practical estimates of vocabulary size
  dependent on word definition, the degree of language input and the
  participant’s age.
\newblock {\em Frontiers in Psychology}, 7, 2016.

\bibitem{deberta}
Pengcheng He, Xiaodong Liu, Jianfeng Gao, and Weizhu Chen.
\newblock Deberta: Decoding-enhanced {BERT} with disentangled attention.
\newblock {\em CoRR}, abs/2006.03654, 2020.

\bibitem{sennrich-etal-2016-neural}
Rico Sennrich, Barry Haddow, and Alexandra Birch.
\newblock Neural machine translation of rare words with subword units.
\newblock In {\em Proceedings of the 54th Annual Meeting of the Association for
  Computational Linguistics (Volume 1: Long Papers)}, pages 1715--1725, Berlin,
  Germany, August 2016. Association for Computational Linguistics.

\bibitem{wordpiece}
Mike Schuster and Kaisuke Nakajima.
\newblock Japanese and korean voice search.
\newblock In {\em 2012 IEEE International Conference on Acoustics, Speech and
  Signal Processing (ICASSP)}, pages 5149--5152, 2012.

\bibitem{kudo-2018-subword}
Taku Kudo.
\newblock Subword regularization: Improving neural network translation models
  with multiple subword candidates.
\newblock In {\em Proceedings of the 56th Annual Meeting of the Association for
  Computational Linguistics (Volume 1: Long Papers)}, pages 66--75, Melbourne,
  Australia, July 2018. Association for Computational Linguistics.

\bibitem{DBLP:conf/iclr/LanCGGSS20}
Zhenzhong Lan, Mingda Chen, Sebastian Goodman, Kevin Gimpel, Piyush Sharma, and
  Radu Soricut.
\newblock {ALBERT:} {A} lite {BERT} for self-supervised learning of language
  representations.
\newblock In {\em 8th International Conference on Learning Representations,
  {ICLR} 2020, Addis Ababa, Ethiopia, April 26-30, 2020}. OpenReview.net, 2020.

\bibitem{DBLP:conf/emnlp/KudoR18}
Taku Kudo and John Richardson.
\newblock Sentencepiece: {A} simple and language independent subword tokenizer
  and detokenizer for neural text processing.
\newblock In Eduardo Blanco and Wei Lu, editors, {\em Proceedings of the 2018
  Conference on Empirical Methods in Natural Language Processing, {EMNLP} 2018:
  System Demonstrations, Brussels, Belgium, October 31 - November 4, 2018},
  pages 66--71. Association for Computational Linguistics, 2018.

\bibitem{porter2001snowball}
Martin~F Porter.
\newblock Snowball: A language for stemming algorithms, 2001.

\bibitem{wolf2019huggingface}
Thomas Wolf, Lysandre Debut, Victor Sanh, Julien Chaumond, Clement Delangue,
  Anthony Moi, Pierric Cistac, Tim Rault, R{\'{e}}mi Louf, Morgan Funtowicz,
  and Jamie Brew.
\newblock Huggingface's transformers: State-of-the-art natural language
  processing.
\newblock {\em CoRR}, abs/1910.03771, 2019.

\end{thebibliography}

\end{document}